\documentclass[10pt,twocolumn,letterpaper]{article}

\usepackage[pagenumbers]{cvpr} 

\usepackage[accsupp]{axessibility}  

\usepackage{multirow}

\usepackage{algpseudocode}
\usepackage{algorithm}
\usepackage{tabularx}
\usepackage{amsmath}
\usepackage{tabularray}
\usepackage{xcolor}
\definecolor{darkgreen}{RGB}{34,139,34}

\usepackage{subcaption} 
\usepackage{listings}
\usepackage{xcolor}
\usepackage{enumitem}
\makeatletter
\newcommand{\multiline}[1]{%
  \begin{tabularx}{\dimexpr\linewidth-\ALG@thistlm}[t]{@{}X@{}}
    #1
  \end{tabularx}
}
\makeatother

\definecolor{cvprblue}{rgb}{0.21,0.49,0.74}
\usepackage[pagebackref,breaklinks,colorlinks,allcolors=cvprblue]{hyperref}

\title{From Failures to Fixes: LLM-Driven Scenario Repair for Self-Evolving Autonomous Driving}

\author{
Xinyu Xia$^{1}$,
Xingjun Ma$^{2}$,
Yunfeng Hu$^{1}$,
Ting Qu$^{1}$,\\
Hong Chen$^{3}$,
Xun Gong$^{1,\dagger}$ \\
\\
$^1$Jilin University \quad
$^2$Fudan University \quad
$^3$Tongji University \\
\\
$^\dagger$Corresponding author
}
\begin{document}
\maketitle

\begin{abstract}
Ensuring robust and generalizable autonomous driving requires not only broad scenario coverage but also efficient repair of failure cases, particularly those related to challenging and safety-critical scenarios. However, existing scenario generation and selection methods often lack adaptivity and semantic relevance, limiting their impact on performance improvement. In this paper, we propose \textbf{SERA}, an LLM-powered framework that enables autonomous driving systems to self-evolve by repairing failure cases through targeted scenario recommendation. By analyzing performance logs, SERA identifies failure patterns and dynamically retrieves semantically aligned scenarios from a structured bank. An LLM-based reflection mechanism further refines these recommendations to maximize relevance and diversity. The selected scenarios are used for few-shot fine-tuning, enabling targeted adaptation with minimal data. Experiments on the benchmark show that SERA consistently improves key metrics across multiple autonomous driving baselines, demonstrating its effectiveness and generalizability under safety-critical conditions.
\end{abstract}

\section{Introduction}
\label{sec:intro}

\begin{figure*}[t]
  \centering
   \includegraphics[width=0.92\linewidth]{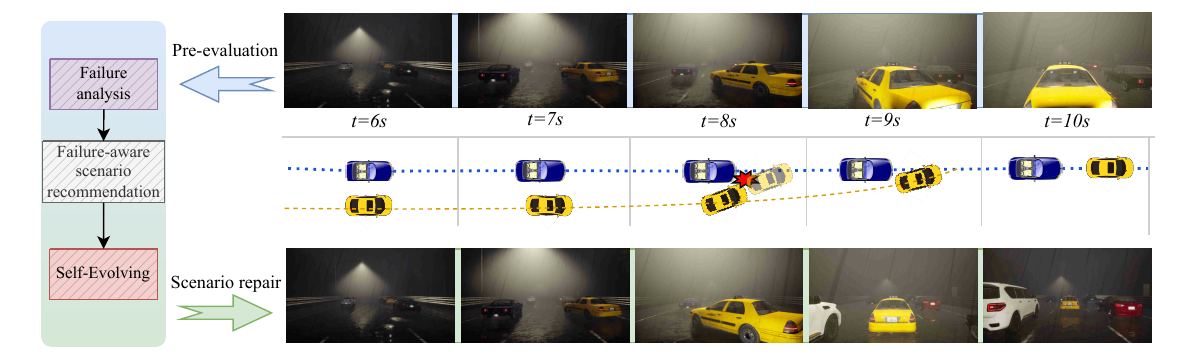}
   \caption{
   Conceptual illustration of \textbf{SERA}. The system performs pre-evaluation to detect failure cases, leverages failure-aware scenario recommendation to retrieve vulnerable instances, and applies self-evolving scenario repair for targeted model adaptation. An example on the right shows a failure due to low-visibility collision that is successfully repaired through efficient fine-tuning, leading to improved decision-making under safety-critical conditions.
   }
   \label{fig:teaser}
   \vspace{-1ex}
\end{figure*}

Autonomous driving technologies have achieved significant advancements in recent years, promising substantial enhancements in transportation safety, efficiency, and reliability~\cite{chen2024end, omeiza2021explanations, huang2020autonomous, li2023survey}. Modern autonomous driving systems rely on sophisticated learning-based algorithms spanning perception, decision-making, control, and coordination. Despite remarkable progress, ensuring consistent and robust performance under complex and dynamically evolving real-world conditions remains challenging~\cite{ren2022survey,10105457}. A key obstacle arises from inadequate exposure of these systems to rare but critical driving situations—such as pedestrian occlusions, intricate merges, or adverse weather—which are significantly underrepresented in typical training datasets~\cite{gehrig2021dsec, ding2023survey}. Consequently, these under-addressed scenarios severely limit system reliability and generalization.

Traditionally, driving scenarios are manually constructed using historical crash records, regulatory guidelines, or scripted simulations~\cite{ulbrich2015defining, wang2022autonomous}. While intuitive and interpretable, these manual approaches suffer critical limitations: they are labor-intensive, challenging to scale effectively, and inherently biased toward previously known or encountered conditions. More importantly, handcrafted scenario generation techniques often miss subtle yet impactful failures related to ambiguous pedestrian intentions, minor visibility variations, or complex interactions among traffic agents. To overcome these issues, automated scenario retrieval techniques leveraging extensive databases have emerged~\cite{hornauer2019driving, wang2024rac3, chang2022metascenario}. However, these methods usually depend on rigid scenario descriptors and handcrafted similarity metrics, thereby struggling to capture the nuanced semantic context necessary for addressing performance-critical failures effectively.

Recent developments in large language models (LLMs) have introduced significant opportunities for semantic scenario understanding, retrieval, and adaptation, owing to their exceptional capabilities in contextual reasoning and high-level abstraction~\cite{deng2023target, xi2025rise, tian2024llm, cai2025text2scenario, tian2024enhancing}. Nevertheless, existing LLM-based research primarily emphasizes scenario generation or interactive scenario creation, with limited exploration of systematically employing LLMs to analyze autonomous driving pre-evaluation outcomes, reason about failures, and recommend efficient scenario repairs. Therefore, there is a critical research gap regarding performance-oriented scenario analysis and efficient repair, particularly by exploiting the reasoning capabilities of LLMs.

To bridge this critical gap, we propose SERA, an innovative framework for \textit{Failure-Aware Scenario Recommendation} and \textit{Self-Evolving Scenario Repair} in autonomous driving. Specifically, SERA systematically analyzes pre-evaluation logs to pinpoint recurring failure patterns and dynamically retrieves semantically aligned scenarios from a structured scenario bank via Failure-Aware Scenario Recommendation. These initial recommendations undergo further semantic refinement through an LLM-powered reflection mechanism to ensure optimal relevance and diversity. The selected scenarios are subsequently used in a Self-Evolving Scenario Repair process, enabling the autonomous driving model to efficiently adapt and effectively repair its identified performance shortcomings. By combining scenario recommendation and targeted model adaptation into a coherent closed-loop system, SERA facilitates continuous improvement, significantly enhancing the robustness and generalization capabilities of autonomous driving systems under diverse and challenging conditions.

An overview of the proposed SERA framework is illustrated in Figure~\ref{fig:teaser}, highlighting the failure-aware scenario recommendation, self-evolving scenario repair process, and an example of autonomous vehicle behavior improvement after scenario refinement. The main contributions of this work are as follows:
\begin{itemize}
    \item We propose SERA, a novel framework that integrates Failure-Aware Scenario Recommendation and Self-Evolving Scenario Repair to systematically enhance the robustness of autonomous driving systems.
    
    \item SERA introduces a unified pipeline that performs pre-evaluation failure analysis, semantic-driven scenario retrieval, and reflection-guided refinement, enabling adaptive and safety-critical model improvement.
    
    \item Extensive experiments on benchmark datasets demonstrate that SERA significantly outperforms state-of-the-art baselines, especially under diverse and challenging driving conditions.
\end{itemize}

\section{Related Work}

\subsection{End-to-End Autonomous Driving}
The development of autonomous driving systems has historically progressed through two major paradigms. Traditional pipelines adopt a modular architecture, decomposing the task into distinct perception~\cite{carion2020end,li2024bevformer,wang2022detr3d}, prediction~\cite{mao2023leapfrog,yao2024trajclip}, and planning~\cite{hu2021safe,teng2023motion} modules. Although this structure facilitates interpretability and enables independent component optimization, it often suffers from error propagation across modules and lacks holistic optimization toward final driving objectives.
In contrast, end-to-end (E2E) autonomous driving frameworks~\cite{chen2024vadv2,hu2023planning,jiang2023vad} aim to overcome these limitations by jointly optimizing perception, prediction, and planning within a unified learning system. These approaches directly map raw sensor inputs to driving actions, thereby improving adaptability to complex and diverse driving scenarios. Recent E2E models demonstrate strong potential in achieving more robust and globally consistent behavior, although challenges remain in ensuring interpretability, safety guarantees, and generalization under long-tail real-world conditions.

\subsection{Scenario-Based Testing for Autonomous Driving}

Scenario-based testing is fundamental for ensuring autonomous driving robustness under diverse, real-world conditions~\cite{pathrudkar2023safr, ulbrich2015defining}. Early approaches primarily relied on manually defined templates or historical crash data, limiting scalability and the discovery of rare, unexpected failures~\cite{fremont2020formal}. To enhance coverage, recent works introduced automated methods, leveraging simulation-based complexity assessment~\cite{ghodsi2021generating}, genetic algorithms~\cite{zhou2022genetic}, and adversarial reinforcement learning~\cite{chen2021adversarial}. However, these techniques typically require explicitly defined seed scenarios or optimization targets, lacking adaptive mechanisms that dynamically respond to pre-evaluation results.

\subsection{Few-shot Learning in Autonomous Driving}

Few-shot learning has been increasingly explored to enhance data efficiency and model generalization in autonomous driving tasks, especially when acquiring large-scale annotated data is costly~\cite{wang2020frustratingly, zhang2022meta}. Recent methods have employed fine-tuning and meta-learning strategies, demonstrating promising performance gains with minimal training examples~\cite{wang2020frustratingly, duan2017one}. Few-shot methodologies have also been leveraged to optimize scenario selection for targeted autonomous driving evaluation~\cite{li2024few}. Nevertheless, existing approaches typically overlook semantic alignment between scenario selection and specific failure cases identified during pre-evaluation.

\begin{figure*}[t]
    \centering
    \includegraphics[width=\textwidth]{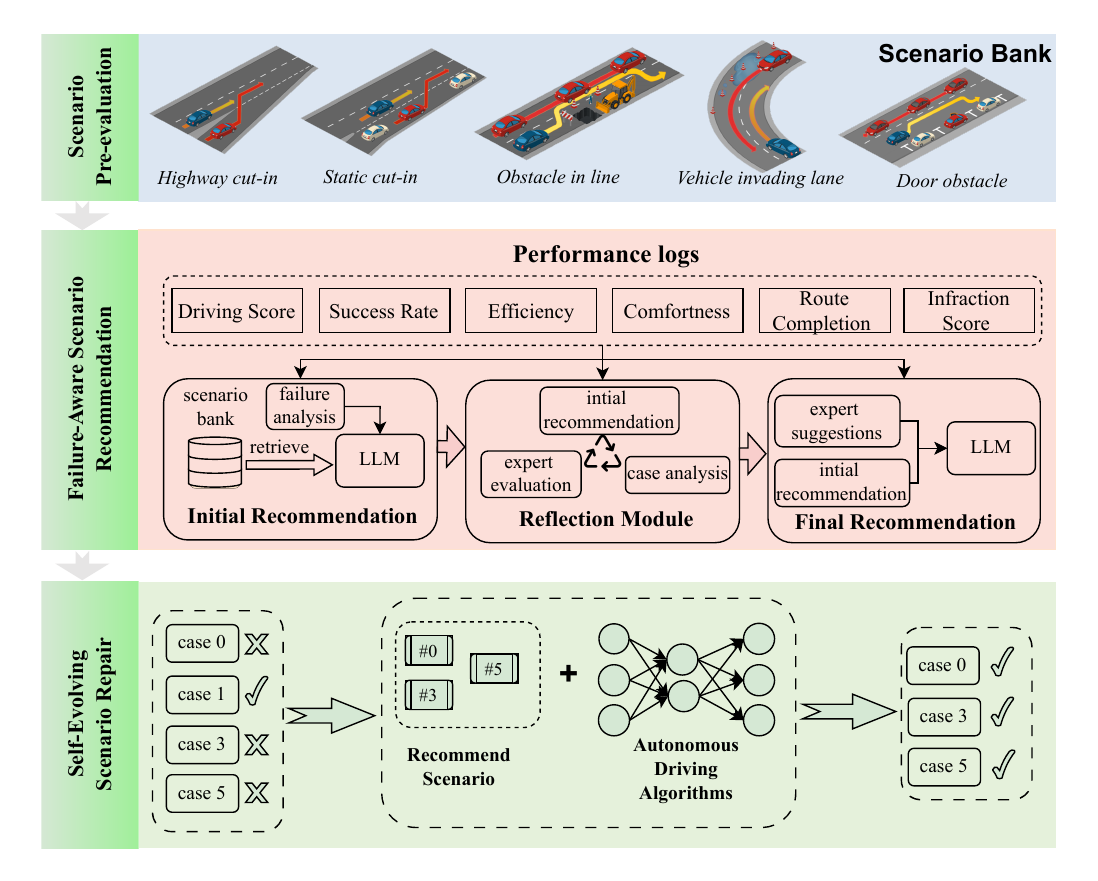}
    \caption{Overview of the proposed SERA framework.}
    \label{framework}
\end{figure*}

\subsection{Large Language Models for Scenario Understanding and Generation}

Recent advancements in large language models (LLMs) have enhanced scenario generation and semantic understanding capabilities in autonomous driving~\cite{zhong2023language, li2023scenarionet, deng2023target}. Current research typically uses LLMs to guide scenario creation with improved semantic fidelity or translate high-level descriptions and traffic rules into simulations~\cite{miceli2023dialogue, cao2023robot}. However, existing LLM-based approaches primarily focus on scenario synthesis or natural language interfaces, neglecting the analysis of pre-evaluation results and efficient, adaptive scenario recommendation.

Despite significant progress in scenario-based testing, few-shot learning, and the application of LLMs, existing methods generally overlook the integration of failure analysis, semantic scenario retrieval, and adaptive recommendation based on pre-evaluation results. In contrast, our proposed SERA framework addresses these limitations by systematically combining these critical aspects, thereby enabling efficient scenario repair and self-evolving adaptation for autonomous driving systems.

\section{Methodology}

\subsection{Overview}
As illustrated in Figure~\ref{framework}, we propose the SERA framework, a safety-oriented self-evolving mechanism designed to systematically analyze, understand, and repair failures in autonomous driving systems through efficient scenario recommendation and adaptation. Different from conventional training methods that broadly focus on performance optimization, our approach explicitly addresses critical safety concerns by identifying and rectifying specific failure scenarios detected during pre-evaluation. 

Specifically, given the performance logs generated from a rigorous pre-evaluation process, SERA utilizes a large language model (LLM) to analyze and interpret failure patterns. These insights guide a \textit{Failure-Aware Scenario Recommendation} pipeline that retrieves semantically relevant scenarios from a structured Scenario Bank. The recommended scenarios are further refined by an LLM-based reflection module to ensure semantic alignment and diversity, producing high-quality, efficient scenarios for subsequent \textit{Self-Evolving Scenario Repair}. This iterative repair process significantly enhances the robustness and safety of autonomous driving systems, particularly under challenging, rare, and safety-critical conditions. The overall process and algorithmic details of our proposed SERA framework are summarized in Algorithm~\ref{alg:sera}.
For clarity, we summarize the key notations and their corresponding descriptions used throughout our framework in Table~\ref{tab:notations}.
\begin{table}[h]
\centering
\caption{Summary of Notations}
\label{tab:notations}
\begin{tabular}{c p{7cm}} 
\hline
\textbf{Symbol} & \textbf{Description} \\
\hline
$S$ & \parbox[t]{6cm}{Structured driving scenario (weather, time, location, scene context)} \\
$T_S$ & \parbox[t]{6cm}{Textual description generated from $S$ via Scenario Descriptor} \\
$\mathcal{B}$ & \parbox[t]{6cm}{Scenario bank containing textual scenarios} \\
$o$ & \parbox[t]{6cm}{Agent observation (sensor input at a time step)} \\
$a$ & \parbox[t]{6cm}{Agent action (trajectory output or control command)} \\
$\pi_{\theta}$ & \parbox[t]{6cm}{Autonomous driving policy parameterized by $\theta$} \\
$\tau$ & \parbox[t]{6cm}{Pre-evaluation driving route} \\
$\mathcal{T}$ & \parbox[t]{6cm}{Set of all pre-evaluation routes} \\
$\ell(\tau, \pi_{\theta})$ & \parbox[t]{6cm}{Performance log collected on route $\tau$} \\
$\mathcal{L}$ & \parbox[t]{6cm}{Set of all performance logs} \\
$p$ & \parbox[t]{6cm}{Individual failure pattern extracted from logs} \\
$\mathcal{P}$ & \parbox[t]{6cm}{Set of extracted failure patterns} \\
$r(s, \mathcal{P})$ & \parbox[t]{6cm}{Initial relevance score between scenario $s$ and failure patterns $\mathcal{P}$} \\
$\mathcal{C}$ & \parbox[t]{6cm}{Initial candidate scenario set retrieved from $\mathcal{B}$} \\
$\mathcal{R}$ & \parbox[t]{6cm}{Reflection-generated improvement suggestions} \\
$r'(s, \mathcal{P})$ & \parbox[t]{6cm}{Reflection-enhanced relevance score} \\
$\mathcal{C}'$ & \parbox[t]{6cm}{Final refined scenario set after reflection} \\
\hline
\end{tabular}
\end{table}

\subsection{Scenario Descriptor}

To enable semantic-level reasoning over autonomous driving scenarios, we introduce a Scenario Descriptor that transforms structured, low-level scenario metadata into detailed textual narratives interpretable by large language models (LLMs). Unlike traditional vision datasets where objects and interactions are explicitly labeled, autonomous driving datasets primarily contain environmental configurations and agent trajectories without fine-grained semantic annotations. Consequently, directly leveraging such data for failure analysis and targeted scenario retrieval poses a significant challenge.

Formally, each scenario is represented by a set of core semantic attributes:
\begin{equation}
    S = \{ \phi_{\text{weather}}, \phi_{\text{time}}, \phi_{\text{location}}, \phi_{\text{scene}} \}
\end{equation}
where:
\begin{itemize}[topsep=2pt, itemsep=2pt]
    \item $\phi_{\text{weather}}$ describes environmental conditions (e.g., clear, rainy, foggy);
    \item $\phi_{\text{time}}$ specifies temporal context (e.g., daytime, nighttime, dawn);
    \item $\phi_{\text{location}}$ indicates geographical settings (e.g., urban intersections, highways, roundabouts);
    \item $\phi_{\text{scene}}$ captures scene semantics, including dynamic agent behaviors and static infrastructure.
\end{itemize}

The Scenario Descriptor module $D(\cdot)$ synthesizes these structured attributes into a natural language description \(T_S\), enabling effective semantic retrieval:
\begin{equation}
    T_S = D(\phi_{\text{weather}}, \phi_{\text{time}}, \phi_{\text{location}}, \phi_{\text{scene}})
\end{equation}

This textualization process bridges the gap between structured environment metadata and the language-centric reasoning capabilities of LLMs. By generating interpretable and semantically rich descriptions, the Scenario Descriptor ensures that downstream retrieval and recommendation stages can reason about scenario relevance at a high level of abstraction, even in the absence of manually annotated semantic labels.

\subsection{Scenario Bank}

The Scenario Bank serves as a structured and semantically enriched repository for autonomous driving scenarios, designed explicitly to support efficient scenario retrieval following failure analysis. It is fundamentally distinct from the pre-evaluation set $\mathcal{T}$, which is used primarily for broad assessment of autonomous driving performance. While $\mathcal{T}$ aims to reveal system vulnerabilities, the Scenario Bank $\mathcal{B}$ provides a rich source of semantically meaningful scenarios intended for targeted repair.

Formally, the Scenario Bank is defined as:
\begin{equation}
    \mathcal{B} = \{ T_{S_1}, T_{S_2}, \dots, T_{S_N} \}
\end{equation}
where each $T_{S_i}$ represents a textual description of a scenario generated by the Scenario Descriptor.

Each entry in $\mathcal{B}$ captures essential semantic properties, including the driving task (e.g., merging, intersection handling), environmental conditions (e.g., rain, fog, nighttime driving), and potential risk factors (e.g., occluded pedestrians, aggressive cut-ins). These descriptions are constructed to reflect both explicit and latent factors affecting autonomous driving behavior, ensuring that the retrieval process can match subtle failure patterns revealed during pre-evaluation.

\subsection{Failure-Aware Scenario Recommendation}
Conventional methods typically select scenarios based solely on training losses or generic performance metrics, lacking the semantic reasoning capability needed to pinpoint the nuanced contexts underlying safety-critical failures. To address this limitation, we propose a Failure-Aware Scenario Recommendation module that explicitly leverages LLM reasoning over detailed performance logs rather than merely numerical metrics.

\subsubsection{Performance Log Analysis}

To enable failure-aware scenario repair, we first conduct pre-evaluation by executing the autonomous driving policy $\pi_{\theta}$ across a set of designated test routes $\mathcal{T}$. During each route $\tau \in \mathcal{T}$, the agent observes the environment and produces an action sequence, forming an interaction trajectory.

Formally, for each $\tau$, the trajectory $\xi_{\tau}$ is defined as:
\begin{equation}
    \xi_{\tau} = \{(o_t, a_t) \mid a_t = \pi_{\theta}(o_t), \, t = 0, \dots, T\}
\end{equation}
where $o_t$ denotes the observation at time step $t$, and $T$ is the episode horizon.

Based on the trajectory $\xi_{\tau}$, we generate the performance log as:
\begin{equation}
    \ell(\tau, \pi_{\theta}) = \mathcal{M}(\xi_{\tau})
\end{equation}
where $\mathcal{M}(\cdot)$ is a structured evaluation mapping that records semantic failures, including route deviations, collisions, traffic violations, and other safety-critical infractions.

Collectively, the pre-evaluation process produces a set of performance logs:
\begin{equation}
    \mathcal{L} = \{ \ell(\tau_i, \pi_{\theta}) \mid \tau_i \in \mathcal{T} \}
\end{equation}

Unlike conventional scalar loss functions, performance logs $\mathcal{L}$ provide fine-grained semantic insights into agent behavior under real-world conditions. For instance, a log entry may explicitly document events such as \textit{"Agent ran a red light at (x=341.25, y=209.1, z=0.104)"} or \textit{"Agent deviated from the route at (x=95.92, y=165.673, z=0.138)"}, with precise spatiotemporal metadata. These detailed descriptions form the foundation for downstream failure pattern extraction and targeted scenario retrieval.

\subsubsection{Initial Scenario Recommendation}

Upon obtaining the pre-evaluation performance logs $\mathcal{L}$, we employ a large language model (LLM) to perform semantic reasoning and extract a set of failure patterns $\mathcal{P}$. Each failure pattern $p \in \mathcal{P}$ characterizes a distinct safety-critical weakness identified in the driving policy $\pi_{\theta}$, such as improper lane merging under rain or delayed braking near intersections.

Formally, the extraction process is expressed as:
\begin{equation}
    \mathcal{P} = \text{AnalyzeLLM}(\mathcal{L})
\end{equation}
where $\text{AnalyzeLLM}(\cdot)$ denotes the LLM-based semantic interpretation of failure causes from the performance logs.

To recommend scenarios relevant to these failures, we define a relevance score $r(s, \mathcal{P})$ between each scenario $s \in \mathcal{B}$ in the Scenario Bank and the extracted failure patterns:
\begin{equation}
    r(s, \mathcal{P}) = \frac{1}{|\mathcal{P}|} \sum_{p \in \mathcal{P}} \phi(T_s, p)
\end{equation}
where $\phi(T_s, p)$ measures the semantic similarity between the scenario description $T_s$ and a failure pattern $p$, implemented via embedding-based similarity or LLM-driven reasoning.

The initial candidate set $\mathcal{C}$ is then obtained by selecting the top-$K$ scenarios with the highest cumulative relevance:
\begin{equation}
    \mathcal{C} = \underset{\substack{\mathcal{C} \subseteq \mathcal{B} \\ |\mathcal{C}| = K}}{\arg\max} \sum_{s \in \mathcal{C}} r(s, \mathcal{P})
\end{equation}

This retrieval ensures that the candidate scenarios $\mathcal{C}$ semantically align with the agent's failure behaviors, providing a focused basis for subsequent scenario refinement and self-evolving adaptation.

\subsubsection{LLM-based Reflection}

To further enhance the quality of the initially recommended scenarios $\mathcal{C}$, we introduce an \textit{LLM-based Reflection} module. This module performs expert-level semantic evaluation by jointly analyzing the initial scenario candidates $\mathcal{C}$ and the extracted failure patterns $\mathcal{P}$.

Specifically, the reflection process leverages the reasoning capabilities of LLMs to assess the coverage adequacy of $\mathcal{C}$ and diagnose any critical failure aspects that remain insufficiently addressed. Based on this analysis, the reflection module outputs a set of \textit{improvement suggestions} $\mathcal{R}$:
\begin{equation}
    \mathcal{R} = \text{ReflectLLM}(\mathcal{C}, \mathcal{P})
\end{equation}
These suggestions indicate necessary modifications to enhance scenario coverage, semantic diversity, and failure-targeted relevance.

\subsubsection{Final Scenario Refinement}

Based on the improvement suggestions $\mathcal{R}$ generated by the reflection module, we refine the initial scenario set $\mathcal{C}$ to obtain the final selection $\mathcal{C}'$:
\begin{equation}
    \mathcal{C}' = \text{RefineLLM}(\mathcal{C}, \mathcal{R})
\end{equation}
where $\text{RefineLLM}(\cdot)$ denotes an LLM-guided adjustment process.

Specifically, $\mathcal{R}$ provides targeted operations including replacing redundant scenarios, augmenting uncovered failure cases, and prioritizing high-risk contexts. By explicitly incorporating these reflection-driven adjustments, the refined set $\mathcal{C}'$ achieves tighter alignment with critical failure patterns while maintaining semantic diversity, thereby maximizing the effectiveness of subsequent scenario repair.

\subsection{Self-Evolving Scenario Repair}

The final refined scenario set $\mathcal{C}'$ drives the Self-Evolving Scenario Repair phase, explicitly targeting the safety-critical failures identified during pre-evaluation. Unlike traditional fine-tuning strategies that rely purely on minimizing generic training losses, our method leverages semantically aligned scenarios to perform targeted adaptation.

Formally, given the pretrained autonomous driving model $\pi_{\theta}$ with parameters $\theta$, the updated parameters $\theta^*$ are obtained by minimizing the failure-specific risk over $\mathcal{C}'$:
\begin{equation}
    \theta^{*} = \underset{\theta}{\arg\min} \; \mathbb{E}_{(o, a) \sim \mathcal{C}'} \left[ \mathcal{L}_{\text{fail}}(\pi_{\theta}(o), a) \right]
\end{equation}
where $\mathcal{L}_{\text{fail}}$ measures deviation from safe behavior under the curated failure scenarios.

This self-evolving repair mechanism ensures that model updates are directly aligned with critical failure modes, leading to substantial improvements in robustness, safety, and generalization to long-tail driving conditions.

\begin{algorithm}[H]
\caption{SERA: Self-Evolving Scenario Repair for Autonomous Driving}
\label{alg:sera}
\begin{algorithmic}[1]
\State \textbf{Input}: Pre-trained model $\theta$, scenario bank $\mathcal{B}$, pre-evaluation set $\mathcal{T} = \{\tau_1, \tau_2, ..., \tau_M\}$
\State \textbf{Output}: Self-evolved model $\theta^{*}$ after targeted repair
\For{each route $\tau$ in $\mathcal{T}$}
    \State Execute policy $\pi_{\theta}$ on route $\tau$, collect performance log $L_{\tau}$
    \State Extract failure patterns $\mathcal{P}_{\tau}$ by analyzing $L_{\tau}$ with an LLM
    \State Retrieve initial candidate scenarios $\mathcal{C}_{\tau}$ from $\mathcal{B}$ via failure-aware relevance scoring
    \State Generate reflection-driven improvement suggestions $\mathcal{R}_{\tau}$ by LLM-based reflection on $\mathcal{C}_{\tau}$ and $\mathcal{P}_{\tau}$
    \State Refine candidate scenarios to obtain final repair set $\mathcal{C}'_{\tau} = \text{RefineLLM}(\mathcal{C}_{\tau}, \mathcal{R}_{\tau})$
    \State Update $\theta$ by minimizing semantic repair loss over $\mathcal{C}'_{\tau}$:
    \[
    \theta \leftarrow \underset{\theta}{\arg\min} \; \mathbb{E}_{(o, a) \sim \mathcal{C}'_{\tau}} \left[ \mathcal{L}_{\text{fail}}(\pi_{\theta}(o), a) \right]
    \]
\EndFor
\State \Return Final repaired model $\theta^{*}$
\end{algorithmic}
\end{algorithm}

\begin{table*}[h]
\centering
\caption{Overall performance comparison of baseline models with and without SERA.}
\label{tab:sera_performance}
\begin{tabular}{lccccc}
\hline
\textbf{Method}  & \textbf{Input} & \textbf{Driving Score} $\uparrow$ & \textbf{Success Rate (\%)} $\uparrow$ & \textbf{Efficiency} $\uparrow$ & \textbf{Comfortness} $\uparrow$ \\
\hline
AD-MLP\cite{zhai2023rethinking} & Ego State & 7.83 & 0.00 & 44.89 & 26.36 \\
AD-MLP + SERA & Ego State & \textbf{8.28} \textcolor{darkgreen}{(+5.75\%)} & \textbf{0.00} (+0) & \textbf{50.66} \textcolor{darkgreen}{(+12.86\%)} & \textbf{30.03} \textcolor{darkgreen}{(+15.71\%)} \\
\hline
UniAD\cite{hu2023planning} & Ego State + 6 Cameras & 33.64 & 8.12 & 98.50 & 42.00 \\
UniAD + SERA & Ego State + 6 Cameras & \textbf{35.10} \textcolor{darkgreen}{(+4.34\%)} & \textbf{9.58} \textcolor{darkgreen}{(+1.46)} & \textbf{102.34} \textcolor{darkgreen}{(+3.90\%)} & \textbf{43.50} \textcolor{darkgreen}{(+3.57\%)} \\
\hline
VAD\cite{jiang2023vad} & Ego State + 6 Cameras & 34.21 & 8.51 & 105.00 & 43.20 \\
VAD + SERA & Ego State + 6 Cameras & \textbf{35.64} \textcolor{darkgreen}{(+4.18\%)} & \textbf{9.89} \textcolor{darkgreen}{(+1.38)} & \textbf{109.20} \textcolor{darkgreen}{(+4.00\%)} & \textbf{44.70} \textcolor{darkgreen}{(+3.47\%)} \\
\hline
TCP \cite{wu2022trajectory} & Ego State + Front Cameras & 31.78 & 22.34 & 76.50 & 18.00 \\
TCP + SERA & Ego State + Front Cameras & \textbf{33.03} \textcolor{darkgreen}{(+3.93\%)} & \textbf{23.92} \textcolor{darkgreen}{(+1.58)} & \textbf{79.51} \textcolor{darkgreen}{(+3.93\%)} & \textbf{19.50} \textcolor{darkgreen}{(+8.33\%)} \\
\hline
\end{tabular}
\end{table*}

\section{Experiments}

\subsection{Experimental Datasets and Baselines}

We conduct our experiments using Bench2Drive \cite{jia2024bench2drive}, a closed-loop evaluation protocol integrated into the CARLA Leaderboard 2.0, specifically designed for end-to-end autonomous driving (E2E-AD) tasks. The base dataset, comprising 1,000 driving clips, serves as the scenario bank, while performance is evaluated across the official set of 220 benchmark routes.

To comprehensively validate the effectiveness of our proposed SERA, we benchmark it against several state-of-the-art E2E-AD methods:

\begin{itemize}
    \item \textbf{UniAD} \cite{hu2023planning}: A transformer-based method that explicitly models perception and prediction using Transformer Queries to enable effective information flow.
    \item \textbf{VAD} \cite{jiang2023vad}: Leverages a vectorized scene representation and Transformer Queries to improve both inference efficiency and driving performance.
    \item \textbf{AD-MLP} \cite{zhai2023rethinking}: A lightweight baseline that feeds historical ego-vehicle states into a Multi-Layer Perceptron (MLP) for trajectory prediction.
    \item \textbf{TCP} \cite{wu2022trajectory}: Fuses front-camera images and ego-vehicle states to jointly predict trajectories and control commands.
\end{itemize}

\subsection{Implementation Details}

We employ LLaMA-3 8B as the base large language model (LLM) within our SERA framework. All experiments are conducted on a computational setup equipped with four NVIDIA RTX 4090 GPUs. For closed-loop evaluation, each autonomous driving model is executed in CARLA across the standard set of 220 benchmark routes. During the fine-tuning stage, scenarios recommended by SERA, selected from the scenario bank, are used to iteratively refine each baseline model. The hyperparameters for fine-tuning are set as follows: a learning rate of 1e-5, a batch size of 2, and a total of 2 training epochs.

\subsection{Evaluation Metrics}

To ensure comprehensive evaluation, we utilize the official CARLA simulator test server under diverse weather conditions. We adopt the standard CARLA Leaderboard metrics:

\begin{itemize}
    \item Driving Score (DS): A composite metric that non-linearly penalizes infractions as the vehicle progresses along the designated route.
    \item Route Completion (RC): The percentage of the total route distance completed.
    \item Infraction Score (IS): Measures the number and severity of driving infractions incurred during navigation.
\end{itemize}

For consistency in local evaluations, we additionally adopt the Bench2Drive benchmark based on CARLA version 0.9.15. Bench2Drive consists of 220 shorter routes (approximately 150 meters each) spanning Town01 through Town15 under diverse weather conditions. Its official evaluation metrics include:

\begin{itemize}
    \item Driving Score: A composite metric similar to the CARLA Leaderboard DS, adapted for short-distance routes.
    \item Success Rate: The percentage of routes successfully completed without encountering critical infractions.
    \item Efficiency: Assesses time and path optimality throughout the route.
    \item Comfortness: Measures driving smoothness and passenger comfort.
\end{itemize}

\subsection{Experimental Results}
\subsubsection{Quantitative Comparison}

Table~\ref{tab:sera_performance} reports the performance comparison between baseline models and their SERA-enhanced versions across four key metrics. Overall, SERA consistently improves robustness, efficiency, and comfort across diverse model architectures.

Specifically, UniAD demonstrates substantial improvements, with Driving Score increasing by +4.34\% and Success Rate by +1.46 points. Efficiency and Comfortness also rise by +3.90\% and +3.57\%, respectively, indicating that failure-driven repair not only enhances goal completion but also optimizes control smoothness. Similarly, VAD achieves +4.18\% Driving Score and +4.00\% Efficiency gains, suggesting that SERA remains effective even for models with strong baseline robustness.

TCP-traj, despite an already high Success Rate, benefits from an +8.33\% improvement in Comfortness, highlighting that scenario repair through SERA improves not just success likelihood but also the qualitative aspects of driving behavior. This reflects the framework’s ability to enrich long-tail conditions impacting ride quality.

\begin{table*}[h]
\centering
\caption{Performance breakdown across different scenarios.}
\label{tab:sera_skill}
\begin{tabular}{l|lllll|l}
\hline
\multirow{2}{*}{\textbf{Method}} & \multicolumn{5}{c|}{\textbf{Ability (\%) ↑}} & \multirow{2}{*}{\textbf{Avg.}} \\
\cline{2-6}
& \textbf{Merging} & \textbf{Overtaking} & \textbf{Emergency Brake} & \textbf{Give Way} & \textbf{Traffic Sign} \\
\hline
AD-MLP\cite{zhai2023rethinking} & 0.00 & 0.00 & 0.00 & 0.00 & 0.00 & 0.000 \\
AD-MLP + SERA & 0.00 & 0.00 & 0.00 & 0.00 & 0.00 & 0.000 \\
\hline
UniAD\cite{hu2023planning} & 9.46 & 12.50 & 20.00 & 30.00 & 23.03 & 18.598 \\
UniAD + SERA & 12.10\textcolor{darkgreen}{(2.64$\uparrow$)} & 14.00\textcolor{darkgreen}{(1.50$\uparrow$)} & 23.30\textcolor{darkgreen}{(3.30$\uparrow$)} & 32.50\textcolor{darkgreen}{(2.50$\uparrow$)} & 25.20\textcolor{darkgreen}{(2.17$\uparrow$)} & 21.820\textcolor{darkgreen}{(3.22$\uparrow$)} \\
\hline
VAD\cite{jiang2023vad} & 0.13 & 17.50 & 14.54 & 30.00 & 25.55 & 17.544 \\
VAD + SERA & 0.50\textcolor{darkgreen}{(0.37$\uparrow$)} & 19.20\textcolor{darkgreen}{(1.70$\uparrow$)} & 17.00\textcolor{darkgreen}{(2.46$\uparrow$)} & 32.40\textcolor{darkgreen}{(2.40$\uparrow$)} & 27.90\textcolor{darkgreen}{(2.35$\uparrow$)} & 19.800\textcolor{darkgreen}{(2.26$\uparrow$)} \\
\hline
TCP\cite{wu2022trajectory} & 24.29 & 15.00 & 29.09 & 50.00 & 51.67 & 34.810 \\
TCP + SERA & 27.20\textcolor{darkgreen}{(2.91$\uparrow$)} & 16.60\textcolor{darkgreen}{(1.60$\uparrow$)} & 32.10\textcolor{darkgreen}{(3.01$\uparrow$)} & 52.80\textcolor{darkgreen}{(2.80$\uparrow$)} & 54.20\textcolor{darkgreen}{(2.53$\uparrow$)} & 36.980\textcolor{darkgreen}{(2.17$\uparrow$)} \\
\hline
\end{tabular}
\end{table*}

For AD-MLP, although Success Rate remains at 0\%, Efficiency and Comfortness improve by +12.86\% and +14.00\%, respectively. This suggests that while simpler models may struggle with complex scenario repair due to perceptual limitations, they still benefit from targeted fine-tuning in smoother control behavior.

Beyond global metrics, Table~\ref{tab:sera_skill} further evaluates fine-grained driving skills. SERA consistently enhances critical capabilities such as Emergency Braking, Traffic Sign handling, and Merging across major baselines. Notably, UniAD and TCP-traj show marked improvements in Emergency Braking (+3.30\% and +3.01\%), confirming that failure-aware repair mechanisms reinforce safety-critical reactions under challenging scenarios.

These findings collectively demonstrate that SERA not only boosts quantitative driving performance but also systematically strengthens specific competencies essential for reliable and safe autonomous driving under real-world conditions.

\begin{figure}
    \centering
    \includegraphics[width=1\linewidth]{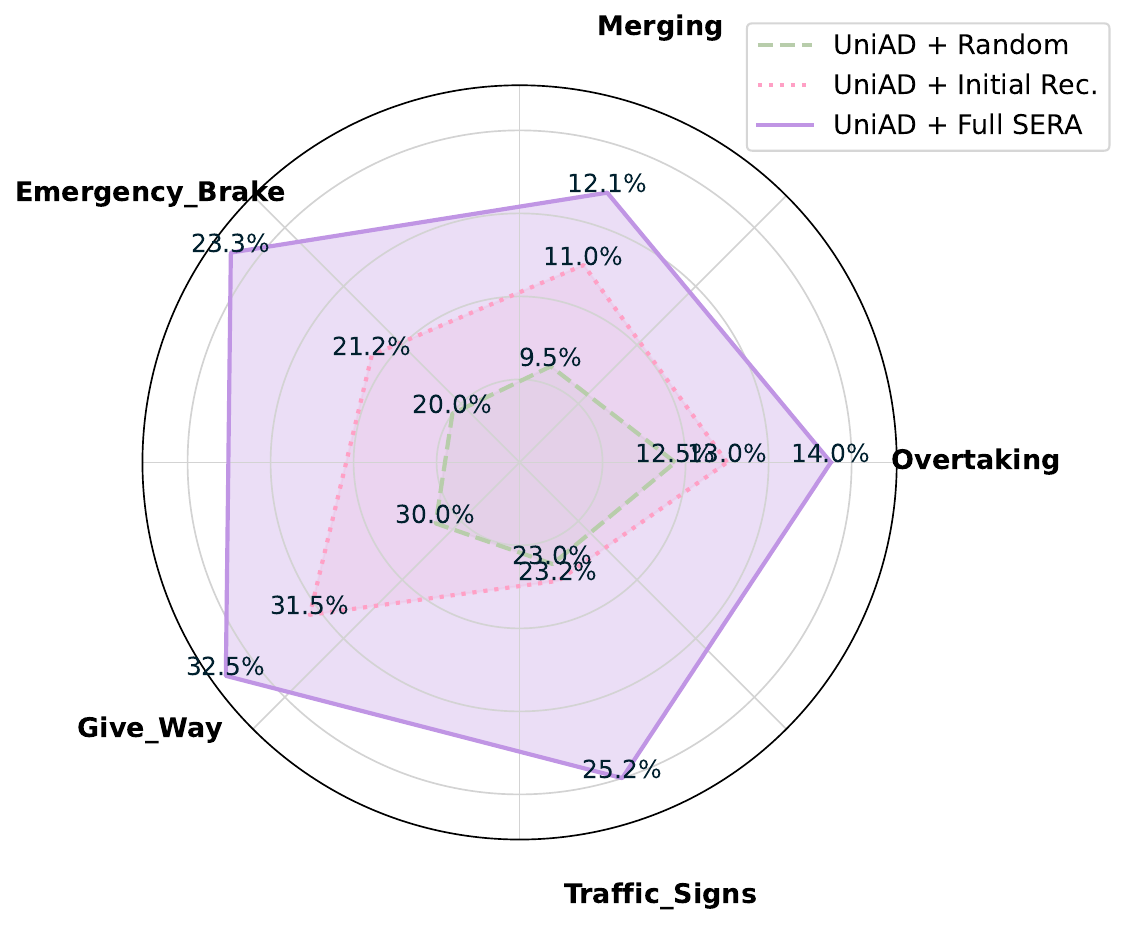}
    \caption{Ability-wise success rate comparison of UniAD under different selection strategies (Random, Initial Rec., and Full SERA). Full SERA consistently improves performance across various driving abilities.}
    \label{fig:ability}
\end{figure}

\begin{figure*}[t]
    \centering
    \includegraphics[width=0.88\linewidth]{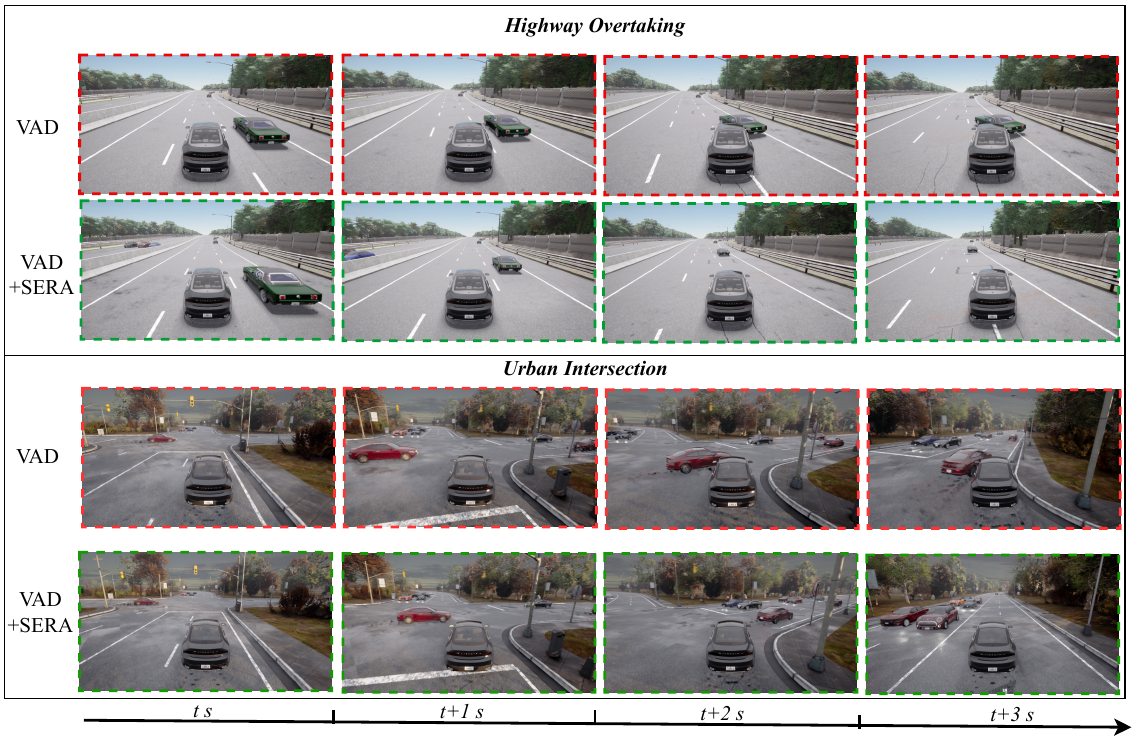}
    \caption{Qualitative comparison between VAD (red dashed borders) and SERA (green dashed borders) across various autonomous driving scenarios. Each column represents a future timestamp (t, t+1s, t+2s, t+3s), showing the behavioral differences between the two methods. SERA demonstrates more consistent and safer navigation compared to VAD.}
    \label{fig:visual}
\end{figure*}

\begin{table}[t]
\centering
\caption{Ablation study of Reflection Module.}
\label{tab:ablation}
\begin{tabular}{lcc}
\hline
\textbf{Method} & \textbf{Driving Score}$\uparrow$ & \textbf{Success Rate}$\uparrow$ \\
\hline
UniAD + Random & 32.20 & 6.85 \\
UniAD + Initial Rec. & 33.25 & 7.52 \\
UniAD + Full SERA & \textbf{35.10} & \textbf{9.58} \\
\hline
VAD + Random & 32.85 & 7.10 \\
VAD + Initial Rec. & 33.90 & 7.90 \\
VAD + Full SERA & \textbf{35.64} & \textbf{9.89} \\
\hline
AD-MLP + Random & 7.10 & 0.00 \\
AD-MLP + Initial Rec. & 7.50 & 0.00 \\
AD-MLP + Full SERA & \textbf{8.28} & \textbf{0.00} \\
\hline
TCP + Random & 29.90 & 20.00 \\
TCP + Initial Rec. & 30.90 & 21.20 \\
TCP + Full SERA & \textbf{33.03} & \textbf{23.92} \\
\hline
\end{tabular}
\end{table}

\subsubsection{Ablation Study}
To assess the contribution of the reflection module in the SERA framework, we conduct an ablation study across all baseline models. As shown in Table~\ref{tab:ablation}, three variants are compared: (i) \textit{Random Selection}, which randomly samples scenarios from the bank without semantic reasoning; (ii) \textit{Initial Recommendation}, which retrieves scenarios based solely on failure pattern relevance; and (iii) \textit{Full SERA}, which incorporates both initial recommendation and reflection-based refinement.

Results show that Random Selection consistently achieves the lowest performance, highlighting that naive scenario augmentation is insufficient for effective model adaptation. For instance, UniAD achieves a Driving Score of only 32.20 and a Success Rate of 6.85\% under random sampling. Initial Recommendation improves upon this baseline, reaching 33.25 and 7.52\% respectively, demonstrating that targeted retrieval based on failure patterns is beneficial.

Notably, Full SERA yields the best performance across all models. With reflection-enhanced refinement, UniAD further improves to a Driving Score of 35.10 and a Success Rate of 9.58\%. Similar trends are observed for VAD and TCP, where reflection consistently provides an additional gain beyond initial retrieval. For simpler models such as AD-MLP, while Success Rate remains unchanged, reflection still leads to measurable improvements in Driving Score, confirming its value even in limited-capacity settings.

These results validate the critical role of the reflection module in the SERA pipeline. By semantically auditing and refining initial recommendations, reflection enhances both the relevance and diversity of selected scenarios, resulting in more effective self-evolving model repair.

To further evaluate the effectiveness of the reflection module across different driving abilities, we present an ability-wise success rate comparison in Figure~\ref{fig:ability}. The radar chart shows that Full SERA consistently achieves superior success rates compared to both Random Selection and Initial Recommendation across all evaluated abilities, including \textit{Overtaking}, \textit{Merging}, \textit{Emergency Brake}, \textit{Give Way}, and \textit{Traffic Signs}. In particular, substantial improvements are observed in complex tasks such as Emergency Brake and Give Way, where Full SERA demonstrates a significant advantage. This comprehensive ability-level analysis further confirms that the reflection mechanism delivers broad and consistent performance gains across diverse and challenging driving scenarios.

\subsubsection{Qualitative Analysis}
To further validate the effectiveness of our proposed method, we present a qualitative comparison between VAD and SERA in Figure~\ref{fig:visual}. Each row shows the predicted future behavior under different driving scenarios, with snapshots taken at successive timestamps ($t$, $t+1$s, $t+2$s, and $t+3$s). Red and green dashed borders represent predictions from VAD and SERA, respectively.

As illustrated, SERA consistently exhibits safer and more stable driving behaviors compared to VAD across diverse environments. In the highway overtaking scenario (top two rows), VAD fails to properly react to the adjacent vehicle, leading to a side collision at $t+3$s. In contrast, SERA maintains a safe longitudinal distance and smoothly avoids the hazard. Similarly, in the intersection crossing scenario (bottom two rows), VAD aggressively proceeds despite the presence of a crossing vehicle, resulting in a collision. Meanwhile, SERA anticipates the dynamic obstacle and successfully yields, preventing a potential accident. 

These observations demonstrate that SERA not only models the multi-agent dynamics more accurately but also generates future plans that are risk-aware and adaptive to complex situations. This qualitative analysis further substantiates the quantitative performance gains reported earlier.

\section{Conclusion}

In this paper, we proposed SERA, a self-evolving scenario repair framework that systematically enhances autonomous driving systems by addressing failure cases through LLM-driven efficient scenario recommendation. Unlike traditional retraining or static scenario generation approaches, SERA leverages large language models to analyze pre-evaluation performance logs, identify critical failure patterns, and recommend semantically aligned scenarios from a structured scenario bank. Through a reflection-driven refinement process, SERA ensures high relevance and diversity in its recommendations, enabling efficient and safety-critical model adaptation. Extensive experiments demonstrate that SERA consistently improves key driving performance metrics across multiple baselines, even under incomplete scenario bank conditions. Our ablation studies further validate the importance of the reflection mechanism in achieving robust and targeted scenario recovery.

\bibliographystyle{ACM-Reference-Format}
\bibliography{main}

\end{document}